\documentclass[11pt]{article}

\usepackage[preprint]{acl}

\usepackage{times}
\usepackage{latexsym}

\usepackage[T1]{fontenc}

\usepackage[utf8]{inputenc}

\usepackage{microtype}

\usepackage{inconsolata}

\usepackage{graphicx}
\usepackage{minted}
\usepackage{booktabs}
\usepackage{placeins}
\usepackage{listings}
\usepackage{bm}
\usepackage{enumitem}

\title{Assessing LLM Reliability on Temporally Recent Open-Domain Questions}

\author{
 \textbf{Pushwitha Krishnappa\textsuperscript{1}},
 \textbf{Amit Das\textsuperscript{2}},
 \textbf{Vinija Jain\textsuperscript{3,4\,\textasteriskcentered}}\thanks{Work does not relate to position at Google.},
 \\
 \textbf{Tathagata Mukherjee\textsuperscript{1}}
 \textbf{Aman Chadha\textsuperscript{3, 5}}\thanks{Work does not relate to position at Apple.}
\\
\\
 \textsuperscript{1}University of Alabama Huntsville,
  \textsuperscript{2}University of North Alabama,\\
 \textsuperscript{3}Stanford University,
 \textsuperscript{4}Google,
 \textsuperscript{5}Apple 
\\
\small{
  \textbf{Corresponding author:} Pushwitha Krishnappa (\href{mailto:email@domain}{pk0055@uah.edu})
 }
}

\begin{document}
\maketitle
\begin{abstract}

Large Language Models (LLMs) are increasingly deployed for open-domain question answering, yet their alignment with human perspectives on temporally recent information remains underexplored. We introduce RECOM (Reddit Evaluation for Correspondence of Models), a benchmark dataset of 15,000 recent Reddit questions from September 2025 paired with community-derived reference answers. We investigate how four open-source LLMs (Llama-3.1-8B, Mistral-7B, Gemma-2-9B, and GPT-OSS-20B) respond to these questions, evaluating alignment using lexical metrics (BLEU, ROUGE), semantic similarity (BERTScore, MoverScore, cosine similarity), and logical inference (NLI). Our central finding is a striking semantic-lexical paradox: all models achieve over 99\% cosine similarity with references despite less than 8\% BLEU-1 overlap - a 90+ percentage point gap indicating that models preserve meaning through extensive paraphrasing rather than lexical reproduction. MoverScore (51–53\%) confirms this pattern, occupying an intermediate position that reflects the optimal transport cost of semantic alignment. Furthermore, model scale does not predict performance: Mistral-7B (7B parameters) outperforms GPT-OSS-20B (20B parameters) across all metrics. NLI analysis reveals that contradiction rates remain below 7\%, suggesting models rarely generate content that directly conflicts with human consensus. These findings challenge the reliability of lexical metrics for evaluating abstractive generation and argue for multi-dimensional evaluation frameworks that capture semantic fidelity beyond surface-level text matching. The RECOM dataset is publicly available - \url{https://anonymous.4open.science/r/recom-D4B0}

\end{abstract}

\section{Introduction}

Large Language Models (LLMs) have demonstrated impressive performance across a wide range of natural language understanding and generation tasks. Their growing adoption in conversational agents, search assistants, and question-answering systems has raised questions about how well these models align with human perspectives, particularly when responding to recent or evolving information. This issue becomes particularly critical when models are applied to recent or rapidly evolving content, where the coverage of training data may be limited.

Most prior work on LLM evaluation focuses on benchmark datasets, static knowledge sources, or synthetic prompts. In contrast, real-world user questions, especially those posed on online platforms such as Reddit, often reflect current events, emerging trends, and evolving community knowledge. These questions present a challenging testbed for evaluating LLM reliability, as correct answers may not be well-established or universally agreed upon.

In this work, we study how LLM-generated answers align with community perspectives on recent Reddit questions from September 2025, leveraging aggregated human responses as reference answers. Instead of relying on a single human-annotated gold answer, we summarize multiple human responses to construct a reference answer that captures diverse perspectives. We then evaluate how closely different LLMs align with this reference when answering the same questions.

Our contributions are as follows:
\begin{enumerate}
    \item We introduce a novel dataset of 11,515 real-world questions paired with community-derived reference responses, where the questions require more than simple factoids.
    \item We compare four open-source LLMs using a comprehensive multi-dimensional evaluation framework spanning lexical metrics (BLEU, ROUGE), semantic similarity (BERTScore, cosine similarity), soft semantic alignment via optimal transport (MoverScore), and logical inference (NLI), revealing a striking semantic-lexical paradox where models achieve 99\%+ semantic similarity despite <8\% lexical overlap.
    \item We demonstrate that model scale does not predict alignment performance, with Mistral-7B (7B parameters) outperforming GPT-OSS-20B (20B parameters) across all metrics.
    \item We show that MoverScore occupies an intermediate position (51–53\%) between lexical and embedding-based metrics, providing a more granular view of semantic alignment.
\end{enumerate}



\begin{figure*}
\centering
  \includegraphics[width=.75\linewidth]{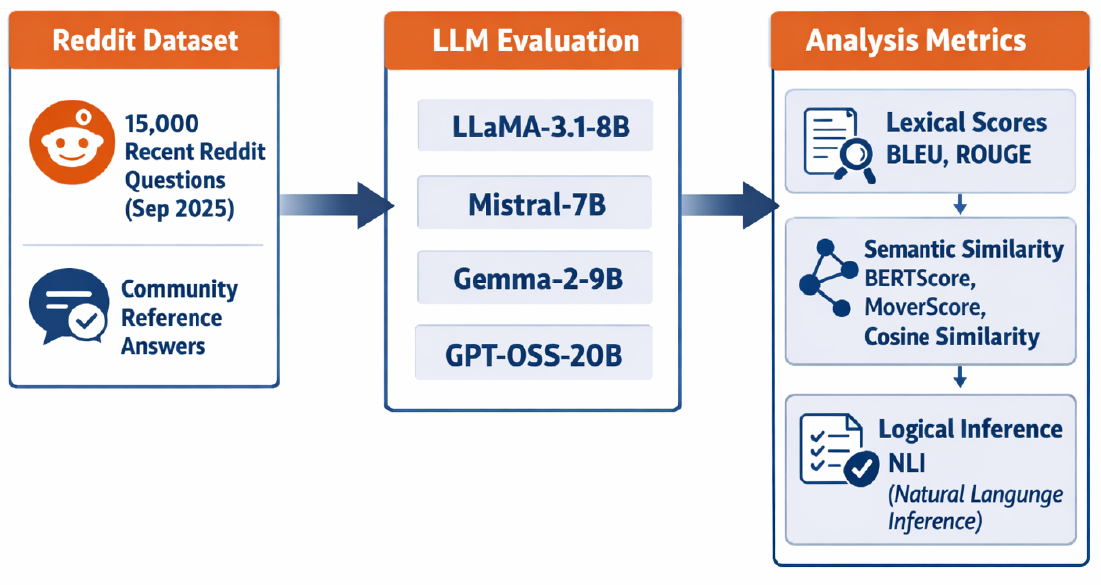} 
  \caption{The LLM Alignment Evaluation for RECOM. 1. Extract Reddit questions and summarize human responses to construct reference answers. 2. Four LLMs generate responses to the same questions. 3. Evaluate outputs using lexical, semantic, and NLI metrics.}
  \label{fig:workflow}
\end{figure*}

\section{Related Work}
The reliability of neural text generation has been extensively studied and evaluated, particularly in tasks such as summarization, machine translation, and question answering. A key concern is the generation of fluent but unsupported or factually incorrect content  \citep{maynez2020faithfulness, ji2023survey}. Prior work distinguishes between intrinsic hallucination, where generated content contradicts available evidence, and extrinsic hallucination, where models introduce unverifiable or fabricated information \citep{dziri2022faithdial}.

Recent research has focused on evaluating factual consistency in LLMs using automatic metrics, entailment-based verification, and prompting strategies. Examples include factual consistency evaluation \citep{kryscinski2020evaluating}, question-answering–based verification methods \citep{honovich2022qafacteval}, and approaches that encourage models to explicitly express uncertainty \citep{lin2022teaching}. However, most of these studies rely on static benchmarks or curated datasets, which limits their applicability to real-world dynamic information needs.

Open-domain question answering (ODQA) systems are typically evaluated on benchmark datasets such as Natural Questions \cite{kwiatkowski2019natural}, TriviaQA \cite{joshi2017triviaqa}, and WebQuestions, which mainly contain well-established factual knowledge. Although these benchmarks allow for standardized evaluation, they do not capture evolving or recently emerging information. Retrieval-based ODQA methods have shown a strong performance on such benchmarks \citep{karpukhin2020dense, lewis2020retrieval}, but their evaluation remains largely disconnected from community-driven questions in the real-world.

Community forums such as Reddit have been explored as alternative data sources for evaluating question answering systems due to their linguistic diversity and topical breadth \citep{zhang2021redditqa, gao2022communityqa}. However, these datasets are less commonly used for systematic alignment evaluation. Our work extends this line of research by measuring how closely LLM responses align with community consensus on recent Reddit questions.



Traditional question answering datasets rely on expert annotators or crowd-workers to create gold-standard answers. However, single-reference annotations often fail to capture ambiguity or legitimate disagreement among humans. Previous work has shown that modeling annotator disagreement and using multiple references can improve evaluation robustness \citep{geva2019towards, pavlick2019disagreement}.


In contrast to previous studies that rely on static benchmarks or synthetic tasks, we evaluated LLMs on recent, real-world questions sourced from online communities. By combining temporally recent data, consensus-based reference construction, and multi-dimensional evaluation across lexical, semantic, and logical inference metrics, our work reveals that models can achieve high semantic alignment while exhibiting minimal lexical overlap with human-generated content.

\section{Methodology}

We investigate how well open-source LLMs align with community-derived answers on temporally recent questions. We address three research questions: 
\begin{enumerate}[nosep]
    \item How closely do model responses align with aggregated human perspectives on post-training-cutoff questions?
    \item Do lexical and semantic metrics provide consistent alignment signals? 
    \item Does model scale correlate with alignment performance?
\end{enumerate}

We operationalize alignment through lexical overlap, semantic similarity, and logical inference metrics.

\subsection{Dataset}

We introduce RECOM (Reddit Evaluation for Correspondence of Models), a benchmark dataset for evaluating LLM alignment with community perspectives on temporally recent questions.

\subsubsection{Data Collection}

We collected 132,728 publicly available posts from the community r/AskReddit on the Reddit platform during September 2025. This subreddit provides open-ended questions with multiple human responses across diverse topics, allowing consensus-based reference construction. The temporal window ensures that the questions reflect recent events and information likely absent from the evaluated models' training corpora.

\subsubsection{Engagement-Based Filtering and Sampling}

To ensure sufficient community input for constructing reliable reference answers, we calculated engagement scores for all 132,728 posts. The engagement score for a post is defined as the total number of comments and nested replies received during the collection period. We retained the top 25,000 highest-engagement posts, reasoning that higher engagement indicates questions that resonated with the community and attracted diverse, substantive responses.

From this filtered set, we randomly sampled 15,000 questions for evaluation. This sampling strategy balances computational tractability with breadth of coverage across question types, while random selection minimizes systematic bias toward particular topics or question styles.

\subsubsection{Construction of Reference Answer }
For each question, we aggregated human responses and generated a reference answer using Llama-3.1-8B-Instruct via Ollama (default parameters) through summarization prompts. The prompt (Table \ref{tab:sum-pt}) elicits the central question and a range of perspectives from the responses. This approach approximates community consensus rather than relying on single-annotator judgment. We acknowledge potential self-alignment bias since Llama-3.1-8B is also an evaluated model. We randomly sampled 10\% of the answers and had them evaluated by one of the authors. 

\begin{table}[h!]
\centering
\begin{minipage}{\columnwidth}
\begin{verbatim}
In this task, you are given a Reddit 
discussion consisting of one original 
post and a collection of replies. 
Your goal is to produce a short, 
integrated summary that captures the 
central question raised in the post 
and the main perspectives or arguments 
expressed in the replies. 
Focus only on the content within the 
thread, avoiding any external context.
\end{verbatim}
\end{minipage}
\caption{Summarization Prompt}
\label{tab:sum-pt}
\end{table}

\subsection{Evaluated Models}

We evaluate four open-source LLMs spanning 7B–20B parameters, selected for scale diversity, availability via Ollama, and representation of different model families. All models used default inference parameters.

Methodological Note: Llama-3.1-8B was used for both reference summarization and evaluation, creating potential self-alignment bias. However, the results are agnostic to any self-alignment bias for Llama-3.1-8B as will be observed through the consistency of the observed outcomes across the different LLMs. 

\subsection{Response Generation}

Each model was prompted with identical instructions (Table \ref{tab:ans-gen}) designed to elicit direct, concise answers:

\begin{table}[h!]
\centering
\begin{minipage}{\columnwidth}
\begin{verbatim}
You are answering this question.
Give a direct answer without 
restating or summarizing the question.
Rules:
- Answer ONLY
- Be clear and concise
- Keep it under 50 words
\end{verbatim}
\end{minipage}
\caption{Answer Generation Prompt}
\label{tab:ans-gen}
\end{table}

The prompt constraints ("under 50 words," "Answer ONLY") were designed to elicit focused responses comparable across models, avoiding verbose padding that could artificially inflate similarity scores. We set a maximum token limit of 1,024 tokens to accommodate longer responses when necessary.

All inference was conducted on a workstation equipped with an AMD Ryzen Threadripper PRO 3995WX processor (64 cores, 128 threads), 1 TB RAM and four NVIDIA RTX A5000 GPU (24GB VRAM), with models served through Ollama's local inference framework.

\subsection{Response Filtering}

To ensure that the evaluation focused on substantive model outputs rather than refusal behaviors, we filtered responses containing patterns indicative of the model declining to answer. We implemented regex-based pattern matching to identify responses containing phrases such as:

\begin{itemize}
    \item Direct AI self-references: "as an AI," "I'm a language model," "as an artificial intelligence"
    \item Capability disclaimers: "I don't have," "I cannot experience," "I lack emotions"
    \item Role disclaimers: "I'm not human," "I am not capable of"
\end{itemize}

After filtering, our final evaluation set comprised 11,515 question-response pairs per model, representing a retention rate of 76.8\% of the 15,000 sampled questions. The filtered questions typically involved personal experiences, emotional responses, or physical sensations that the models appropriately declined to simulate.

\subsection{Evaluation Framework}

We employ a multi-dimensional evaluation framework spanning four complementary perspectives on response alignment. This multi-metric approach enables us to distinguish between surface-level textual similarity and deeper semantic correspondence.

\subsubsection{Lexical Overlap Metrics}

Lexical metrics measure direct word and phrase overlap between generated responses and reference summaries.

\textbf{BLEU:} We compute BLEU-1 through BLEU-4 scores to measure n-gram precision at multiple granularities. BLEU \cite{papineni2002bleu} was originally designed for machine translation evaluation and penalizes outputs that diverge lexically from references, even when semantically equivalent.

\textbf{ROUGE} \cite{lin2004rouge}: We report ROUGE-1, ROUGE-2, and ROUGE-L F1 scores to capture recall-oriented lexical similarity. ROUGE-L specifically measures the longest common subsequence, capturing word ordering patterns beyond bag-of-words overlap.

\subsubsection{Semantic Similarity Metrics}

Semantic metrics assess meaning preservation independent of exact wording.

\textbf{BERTScore:} We compute precision, recall, and F1 using RoBERTa-large embeddings \cite{liu2019roberta}. BERTScore \cite{zhang2019bertscore} performs soft token matching using contextual embeddings, rewarding semantic equivalence even when surface forms differ. This metric occupies a middle ground between pure lexical matching and holistic semantic comparison.

\textbf{MoverScore:} We compute MoverScore using contextualized BERT embeddings following \cite{zhao2019moverscore}. Unlike BERTScore's greedy one-to-one token alignment, MoverScore formulates semantic similarity as an optimal transport problem, computing the minimum cost of transforming the token distribution of the generated text into that of the reference using Earth Mover's Distance (EMD). This allows soft many-to-many alignments, capturing cases where meaning is distributed differently across tokens in the two texts.

\textbf{Cosine Similarity:} We compute cosine similarity between mean-pooled RoBERTa-large embeddings of complete generated answers and reference summaries. This metric captures holistic semantic alignment at the document level, abstracting away from token-level correspondences.

\subsubsection{Logical Inference Metrics}

\textbf{Natural Language Inference (NLI):} \cite{bowman2015large} We classify each (generated answer, reference summary) pair using facebook/bart-large-mnli \cite{nie2020adversarial,lewis2020bart} into three categories: entailment (answer logically follows from summary), contradiction (answer conflicts with summary), or neutral (no clear logical relationship). High contradiction rates may indicate factual inconsistency, while high neutral rates suggest the model generates related but non-derivative content.

\subsection{Statistical Analysis}

All pairwise model comparisons are evaluated using the Wilcoxon signed-rank test \cite{casella2024statistical}, a non-parametric test appropriate for paired samples that does not assume normal (Gaussian) distributions. This is important given the skewed distribution of metrics like BLEU scores. We report effect sizes using Cohen's d \cite{cohen2013statistical} to distinguish statistically significant differences from practically meaningful ones, interpreting |d| < 0.2 as negligible, 0.2–0.5 as small, 0.5–0.8 as medium, and > 0.8 as large effects.

\section{Results}

We evaluated four open-source LLMs on 11,515 Reddit questions after filtering refusal responses. We report results across four evaluation dimensions: lexical overlap, semantic similarity, embedding-based alignment, and logical inference patterns. All pairwise model differences reported below are statistically significant (Wilcoxon signed-rank test, p < 0.001) unless otherwise noted. We additionally report effect sizes (Cohen's d) to contextualize the practical significance of observed differences.

\subsection{Lexical Overlap Metrics}

Table \ref{tab:lexical_overlap} presents the BLEU and ROUGE scores across the models. All models exhibit low lexical overlap with reference summaries, with BLEU-1 scores ranging from 0.57\% to 7.58\% and BLEU-4 scores uniformly below 1\%.

\begin{table*}[!tb]
\centering
\begin{tabular}{lcccc}
\hline
\textbf{Model} & \textbf{BLEU-1} & \textbf{BLEU-4} & \textbf{ROUGE-1 F1} & \textbf{ROUGE-L F1} \\
\hline
Gemma2-9B & 0.57 ± 1.41 & 0.06 ± 0.24 & 9.13 ± 6.57 & 6.89 ± 4.61 \\
GPT-OSS-20B & 3.71 ± 4.02 & 0.29 ± 0.54 & 14.28 ± 7.80 & 9.72 ± 4.95 \\
Llama-3.1-8B & 7.58 ± 5.27 & 0.62 ± 1.85 & 18.81 ± 7.93 & 11.54 ± 4.97 \\
Mistral-7B & 6.17 ± 5.38 & 0.68 ± 1.18 & 19.97 ± 7.83 & 13.16 ± 5.14 \\
\hline
\end{tabular}
\caption{Lexical Overlap Metrics (mean ± std). BLEU-1 ranges from 0.57\% to 7.58\%, indicating minimal lexical copying across all models. High standard deviations relative to means reflect heavily skewed distributions, with most responses achieving near-zero n-gram overlap.}
\label{tab:lexical_overlap}
\end{table*}

All values in percentages. Llama-3.1-8B was also used for reference summarization; scores may reflect self-alignment bias.

The scores reveal a consistent model ranking across all lexical metrics: Mistral-7B and Llama-3.1-8B lead, followed by GPT-OSS-20B, with Gemma2-9B trailing substantially. The difference between top and bottom performers is substantial: Mistral-7B achieves 10.8x higher BLEU-1 than Gemma2-9B (6.17\% vs. 0.57\%), corresponding to a large effect size (Cohen's d = 1.43).

Notably, GPT-OSS-20B's larger parameter count (20B) does not yield higher lexical alignment than the 7-8B models. This finding challenges the assumption that model scale directly translates to improved performance on open-domain questions, at least as measured by lexical overlap metrics.

Threshold Analysis: Only 0.3\% of Gemma2-9B responses exceed 10\% BLEU-1 overlap, compared to 29.0\% for Llama-3.1-8B and 21.7\% for Mistral-7B. This distribution indicates that low lexical overlap is not merely a matter of means but reflects a consistent generation strategy across questions.

\subsection{Semantic Similarity}

Despite low lexical overlap, all models achieve high semantic alignment as measured by BERTScore and cosine similarity (Table \ref{tab:semantic-similarity}). This divergence between lexical and semantic metrics constitutes our central finding.


\begin{table*}[!tb]
\centering
\begin{tabular}{lccccc}
\hline
\textbf{Model} & \textbf{BERTScore P} & \textbf{BERTScore R} & \textbf{BERTScore F1} & \textbf{MoverScore} & \textbf{Cosine Sim.} \\
\hline
Gemma2-9B & 85.55 & 81.18 & 83.29 & 50.92 & 99.20 \\
GPT-OSS-20B & 84.48 & 82.25 & 83.33 & 51.84 & 99.10 \\
Llama-3.1-8B & 84.96 & 83.14 & 84.03 & 52.65 & 99.47 \\
Mistral-7B & 86.20 & 83.51 & 84.83 & 53.42 & 99.51 \\
\hline
\end{tabular}
\caption{Semantic Similarity Metrics with 95\% confidence intervals. Despite low lexical overlap, all models achieve high semantic alignment: cosine similarity exceeds 99\%, BERTScore F1 ranges from 83.3\% to 84.8\%, and MoverScore ranges from 50.9\% to 53.4\%. MoverScore's intermediate values reflect the optimal transport cost of token-level semantic alignment. Also used for reference summarization.}
\label{tab:semantic-similarity}
\end{table*}

All models achieve cosine similarity above 99\%, indicating near-perfect alignment in RoBERTa-large's embedding space. The total range spans only 0.41 percentage points (99.10–99.51\%), a dramatic compression compared to the 13-point range observed in ROUGE-1 scores and the 10x range in BLEU-1.
BERTScore F1 scores cluster within a narrow 1.54 percentage point range (83.29–84.83\%), with Mistral-7B achieving the highest scores. While the BERTScore differences are statistically significant (all pairwise p < 0.001), the effect sizes are smaller than those observed for lexical metrics: Mistral-7B versus Gemma2-9B yields d = 0.91 on BERTScore F1 compared to d = 1.50 on ROUGE-1.

MoverScore, which measures semantic similarity via optimal transport over contextualized embeddings, yields scores in the 50.92–53.42\% range across models. Unlike cosine similarity and BERTScore which cluster near ceiling values, MoverScore exhibits greater dispersion (2.50 percentage points) while preserving the same model ranking: Mistral-7B (53.42\%) > Llama-3.1-8B (52.65\%) > GPT-OSS-20B (51.84\%) > Gemma2-9B (50.92\%). The intermediate magnitude of MoverScore substantially higher than lexical metrics yet lower than BERTScore and cosine similarity - reflects its sensitivity to the cost of aligning token-level semantics rather than measuring pure embedding proximity or n-gram overlap. GPT-OSS-20B exhibits the highest MoverScore variance (std = 3.42) with occasional zero-value outliers, consistent with its higher variance observed in cosine similarity (Table \ref{tab:moverscore-stats}) and suggesting more erratic response patterns compared to the smaller models.

Precision-Recall Patterns: Gemma2-9B exhibits the largest precision-recall gap (85.55\% precision, 81.18\% recall, $\Delta$ = 4.37 points), suggesting conservative generation that achieves high precision but omits some reference content. Mistral-7B shows a more balanced profile (86.20\% precision, 83.51\% recall, $\Delta$ = 2.69 points), indicating more comprehensive coverage of reference semantics.

Threshold Analysis: 46.2\% of Mistral-7B responses exceed 85\% BERTScore F1, compared to only 16.4\% for Gemma2-9B. For cosine similarity, 59.2\% of Mistral-7B responses exceed 99.5\%, versus 11.0\% for Gemma2-9B. These distributions confirm that Mistral-7B's superior mean scores reflect consistent high performance rather than a few outlier successes.

\subsection{Model Ranking and Consistency}

Model rankings remain stable across semantic metrics. Mistral-7B achieves highest scores on 51.7\% of questions (cosine similarity) and 57.5\% (BERTScore F1), with Llama-3.1-8B second. Despite its larger parameter count, GPT-OSS-20B achieves highest alignment on only 8.5\% of questions.

Inter-model correlations on BERTScore F1 are moderate (r = 0.21–0.54), indicating questions difficult for one model are not consistently difficult for others. The highest correlation occurs between Llama-3.1-8B and Mistral-7B (r = 0.54). Across all questions, 42.9\% show high inter-model agreement (BERTScore std < 0.01), while only 0.3\% show substantial disagreement (std > 0.05).

\subsection{Logical Inference Patterns}

Table \ref{tab:nli-results} presents NLI classification results using BART-large-MNLI.

\begin{table}[h]
\centering
\small
\begin{tabular}{lccc}
\toprule
\textbf{Model} & \textbf{Entailment} & \textbf{Contradiction} & \textbf{Neutral} \\
\midrule
Gemma2-9B & 9.04\% & 5.15\% & 85.81\% \\
GPT-OSS-20B & 3.43\% & 6.94\% & 89.39\% \\
Llama-3.1-8B & 4.06\% & 5.53\% & 90.41\% \\
Mistral-7B & 8.71\% & 4.31\% & 86.98\% \\
\bottomrule
\end{tabular}
\caption{NLI classification distribution using BART-large-MNLI. Neutral classifications dominate (86--90\%), indicating models generate semantically related but non-entailing content. Contradiction rates remain below 7\% across all models. Also used for reference summarization.}
\label{tab:nli-results}
\end{table}

The majority of response-reference pairs (86–90\%) receive neutral classifications, indicating that generated answers are topically related to but neither logically entailed by nor contradictory to the reference summaries. This pattern is consistent with abstractive paraphrasing: models generate semantically related content without strictly deriving it from or contradicting the reference.

Entailment Patterns: Gemma2-9B and Mistral-7B show higher entailment rates (9.04\% and 8.71\%) despite their different lexical overlap profiles. This suggests these models more frequently generate responses that logically follow from the reference content, possibly indicating closer adherence to the factual claims in human responses.

Contradiction Analysis: Overall contradiction rates remain below 7\% across all models, suggesting that direct factual conflicts between generated answers and human consensus are relatively rare. However, examining individual responses reveals meaningful variation: (a) 6.1–10.0\% of questions elicit at least some contradiction signal (contradiction score > 0), depending on the model (b) GPT-OSS-20B shows the highest rate of strong contradictions (>50\% contradiction score) at 5.5\% of questions and (c) Mistral-7B shows the lowest rate of strong contradictions at 2.1\%. These strong contradiction cases represent instances where model outputs directly conflict with community-expressed perspectives and warrant further investigation.



\subsection{The Semantic-Lexical Divergence}

Our central finding is a striking divergence between lexical and semantic metrics. Models achieve 99\%+ cosine similarity while maintaining less than 8\% BLEU-1 overlap—a gap of approximately 90 percentage points.

Taking Mistral-7B as an example: its cosine similarity (99.51\%) is 16x higher than its BLEU-1 (6.17\%), BERTScore F1 (84.83\%) is 14x higher, and MoverScore (53.42\%) is 8.7x higher. Effect sizes between Mistral-7B and Gemma2-9B show large effects on ROUGE-1 (d = 1.50) and BERTScore F1 (d = 0.91), but only 0.31\% absolute difference on cosine similarity—suggesting models vary in paraphrasing strategies while achieving comparable semantic comprehension.

\section{Discussion} \label{sec:disscuss}

Our results reveal fundamental tensions in how we evaluate LLM-generated text and what different metrics actually capture about model capabilities. We examine four key patterns that emerge from this multi-dimensional evaluation and discuss their implications for research and practice.

\subsection{The Semantic-Lexical Paradox}

Our most striking finding is the dramatic divergence between lexical and semantic evaluation metrics. All four models achieve 99.1–99.5\% cosine similarity with reference summaries, yet their lexical overlap remains minimal (BLEU-1: 0.57–7.58\%). This 90+ percentage point gap indicates that models capture semantic meaning while using almost entirely different words—effectively performing paraphrase at scale.

This paradox has several important implications. First, it demonstrates that modern LLMs have acquired robust paraphrasing capabilities, transforming reference content into semantically equivalent but lexically distinct expressions. Second, it reveals that high embedding similarity can coexist with near-zero n-gram overlap, challenging the assumption that these metrics measure related constructs.

The semantic metrics themselves form a gradient that illuminates different facets of this paradox. Cosine similarity (99.1–99.5\%) captures holistic document-level alignment in embedding space. BERTScore F1 (83.3–84.8\%) performs greedy token-level alignment, yielding lower scores that reflect the cost of imperfect one-to-one token matching. MoverScore (50.9–53.4\%) sits lower still, as its optimal transport formulation penalizes the cumulative cost of moving semantic mass between token distributions. This gradient—from near-ceiling cosine similarity to moderate MoverScore to minimal BLEU—suggests that the apparent paradox reflects not a binary semantic-lexical divide but a continuum of alignment granularities.

The NLI results add another layer of complexity: despite 99\%+ embedding similarity, only 3.4–9.0\% of responses show entailment, with 86–90\% classified as neutral. This demonstrates that two texts can occupy similar semantic space without one logically following from the other—a key distinction between semantic similarity and logical entailment that has implications for hallucination detection.

\subsection{Model Performance Patterns}

Model performance does not scale with parameter count. Mistral-7B achieves the strongest results across all metrics, followed closely by Llama-3.1-8B. Despite having 20 billion parameters, GPT-OSS-20B underperforms both smaller models on every metric, while Gemma2-9B lags substantially behind on lexical measures despite competitive semantic alignment.

Critically, all models achieve >99\% cosine similarity, indicating they grasp core semantic concepts equally well—the differences emerge in how they express those concepts lexically. This suggests that architectural choices, training data composition, and instruction-tuning strategies matter more than raw parameter count for this task. The finding that 7B models outperform a 20B model challenges simplistic assumptions about scaling laws in abstractive generation contexts.

The inter-model correlation analysis (r = 0.21–0.54 on BERTScore) reveals that questions difficult for one model are not consistently difficult for others. This suggests models have different strengths and weaknesses rather than uniform capability profiles, which has implications for ensemble approaches and model selection.

\subsection{Evaluation Metric Implications}

Each metric family reveals different aspects of generation quality. BLEU and ROUGE measure surface form correspondence and severely penalize paraphrastic generation, systematically underestimating quality for abstractive tasks. MoverScore (50.9–53.4\%) occupies a middle ground via optimal transport, reflecting the cost of redistributing semantic content across different lexical realizations. BERTScore (83–85\%) computes token-level semantic similarity, tolerating lexical variation while anchoring to reference structure. Cosine similarity (99\%+) captures holistic document-level alignment. NLI metrics capture logical relationships; the 86–90\% neutral rates indicate outputs and references are topically related but structurally distinct.

The consistent model ranking across all semantic metrics (Mistral-7B > Llama-3.1-8B > GPT-OSS-20B > Gemma2-9B) suggests they capture a shared underlying construct despite different magnitudes. These findings argue for multi-metric evaluation portfolios rather than single-number summaries.

\subsection{Alignment vs. Factual Correctness}

Our evaluation measures alignment with community consensus rather than factual correctness. The low contradiction rates (2-6\%) suggest models rarely conflict with human perspectives, but high alignment does not guarantee accuracy—models could match popular but incorrect responses. The high neutral NLI rates (86-90\%) confirm most outputs discuss related topics differently rather than supporting or contradicting references. Future work should incorporate external knowledge verification to disentangle these factors.

\section{Conclusion}

We present a large-scale study of LLM alignment with community perspectives on temporally recent questions. By comparing four open-source LLMs against reference answers derived from aggregated human responses, our evaluation reflects realistic information-seeking scenarios using 11,515 Reddit questions from September 2025.

Our central finding is a striking semantic-lexical paradox: models achieve over 99\% cosine similarity with references despite less than 8\% BLEU-1 overlap, demonstrating that high semantic alignment can coexist with minimal lexical reproduction. BERTScore (83–85\%) and MoverScore (51–53\%) reveal a gradient of alignment granularities between these extremes. Additionally, model scale does not predict performance—Mistral-7B outperforms the larger GPT-OSS-20B across all metrics, including MoverScore's optimal transport-based evaluation. These results challenge the reliability of lexical metrics for evaluating abstractive generation and argue for multi-dimensional evaluation frameworks combining lexical, semantic, and logical inference measures.

\section{Ethical Considerations}

This work analyzes publicly available data collected from Reddit, an online platform where users voluntarily post questions and responses. We only use content that is publicly accessible and do not attempt to deanonymize users or infer sensitive personal attributes. To further reduce privacy risks, we do not release raw user identifiers or quote individual posts verbatim in the paper.

The reference answers used in this study are generated by summarizing multiple human responses using a large language model. While this approach approximates community consensus, it may inadvertently amplify dominant viewpoints or overlook minority perspectives. We acknowledge this risk and emphasize that our reference summaries should not be interpreted as authoritative ground truth.

All evaluated LLMs are used strictly for research purposes. Our analysis focuses on measuring response alignment rather than endorsing or deploying any specific model. The findings should not be interpreted as guarantees of correctness for real-world applications.

Finally, as model-generated content may include inaccurate or misleading information, we avoid presenting example outputs that could cause harm or propagate false claims. Our study is intended to support safer and more reliable use of LLMs by identifying their limitations when answering recent questions.


\section{Limitations}

Our study has several limitations. First, alignment is measured using automatic similarity metrics (BLEU, ROUGE, BERTScore, and cosine similarity), which capture surface-level and semantic correspondence but not factual correctness. While our multi-metric approach provides robustness, these metrics cannot replace human judgment, especially for ambiguous or opinion-based questions.

Second, reference answers are generated via LLM-based summarization of human responses rather than manual expert annotation. Although this method approximates community consensus, it may introduce summarization bias and propagate errors present in the original responses. Additionally, using Llama-3.1-8B for both summarization and evaluation creates potential self-alignment bias.

Third, our dataset is limited to r/AskReddit posts from September 2025, a subreddit emphasizing open-ended personal questions, opinions, and advice-seeking. Findings may not generalize to technical question-answering, factual retrieval, or domains with objectively verifiable answers. Reddit communities also vary widely in expertise and reliability.

Fourth, we evaluate four open-source models using a single prompting strategy without retrieval augmentation. Performance may differ under alternative prompts, decoding strategies, or retrieval-based settings. The GPT-OSS-20B model has limited documentation, which may affect reproducibility.

Finally, our metrics measure alignment with human consensus rather than factual correctness. Low similarity scores may reflect valid alternative responses rather than errors, while high similarity could mask agreement with incorrect human responses. Future work should incorporate human evaluation and external knowledge verification to disentangle these factors.


\bibliography{custom}

\appendix

\section{Appendix}
\label{sec:appendix}

\subsection{Dataset Sample}
\label{sec:appendix-dataset}

Table~\ref{tab:dataset-sample} presents representative examples showing the diversity of questions and the nature of community-derived reference answers.

\begin{table*}[t]
\centering
\small
\begin{tabular}{p{4cm}p{6.5cm}p{4.5cm}}
\toprule
\textbf{Reddit Question} & \textbf{Reference Summary} & \textbf{Example Model Response (Mistral-7B)} \\
\midrule
What made you realize you are not as young as you used to be? & 
The main question underlying this collection of stories is: "When do you realize you're getting older?" The key discussion themes revolve around the various ways people experience and acknowledge the passage of time, including physical changes, loss of familiarity with popular culture, feelings of irrelevance among younger generations, and a growing sense of responsibility and adulthood. These shared experiences often come as a surprise to individuals who had previously identified themselves as young or youthful, leading to feelings of discomfort, nostalgia, and acceptance of their advancing age. &
Experiencing physical changes such as aches, tiredness after less exertion, or needing glasses, among other things, might hint that you're not as young as before. \\
\midrule
What’s something that instantly makes you lose respect for someone, no matter how successful or attractive they are? & 
The original post asks what behaviors or actions would instantly lead someone to lose respect for another person, regardless of their success or attractiveness. The replies reveal a range of behaviors that erode respect, including racism, rudeness, cruelty, dishonesty, hypocrisy, and lack of empathy towards others. Many responders emphasize the importance of treating people with kindness and respect, and that actions speak louder than external factors like success or physical appearance. & 
Consistently demonstrating a lack of empathy, integrity, or kindness can make others lose respect, regardless of success or attractiveness. \\
\midrule
What's the best way to learn a new language? & 
"Here is a concise summary of the Reddit discussion:
The original poster asks for advice on learning a new language, sparking a lively debate among commenters. The consensus seems to be that immersion in the target language through various means such as full immersion, watching content in the language, and interacting with native speakers is key to successful language acquisition. However, some commenters also suggest more accessible alternatives, including using apps like Duolingo or joining language exchange groups." &
The best way to learn a new language is through immersion, consistent practice, and structured study. Utilize language learning apps, take classes, converse with native speakers, and engage in daily activities in the target language. \\
\midrule 
Reddit, what is a weird thing that you do? & 
In this Reddit discussion, users shared unusual habits and practices they engage in when feeling stressed or anxious. The thread revolves around a central question of what constitutes a "weird" behavior, with participants revealing a diverse array of quirky customs, from talking to plants to exposing themselves to various textures on the soles of their feet. The conversation also touches on mental health, with some users acknowledging that these behaviors may be coping mechanisms for anxiety or other issues. & 
I enjoy alphabetizing my spice rack in reverse order. It's odd, but brings me a strange sense of satisfaction! \\
\midrule
What’s the first sign you usually notice when your mental health starts declining? & 
"Summary
The main question revolves around identifying early warning signs of mental health decline, with individuals sharing their personal experiences and red flags they notice before things spiral out of control. Common signs include changes in daily routines, unexplained sleep or appetite issues, social withdrawal, loss of interest in activities, and feelings of overwhelm or exhaustion. These subtle signals can serve as a catalyst for self-reflection and intervention to prevent further decline.
Key discussion themes:
* Early warning signs of mental health decline
* Personal experiences with red flags and warning signs
* Changes in daily routines and habits
* Unexplained sleep or appetite issues
* Social withdrawal and isolation
* Loss of interest in activities and hobbies
* Feelings of overwhelm, exhaustion, and burnout" &
The first sign often noticed when mental health starts declining can include feelings of persistent sadness, increased anxiety, loss of interest in activities, changes in sleep patterns, and difficulty concentrating. \\
\bottomrule
\end{tabular}
\caption{Sample questions, reference summaries, and model responses from our dataset. References are LLM-generated summaries of multiple human responses; model responses demonstrate the semantic-lexical paradox where meaning is preserved despite different wording.}
\label{tab:dataset-sample}
\end{table*}

\subsection{Illustrative Examples}
\label{sec:appendix-examples}

We present examples spanning the performance spectrum: typical paradox cases (high semantic, low lexical) and edge cases (extractive responses, variance-induced underperformance).

\subsubsection{Paradox Examples: High Semantic, Low Lexical}

These examples demonstrate our central finding---models achieve near-perfect semantic alignment despite minimal lexical overlap.

\textbf{Example 1: Unsolved Mystery (Gemma2-9B)}

\begin{itemize}[nosep,leftmargin=*]
    \item \textbf{Question:} What's the strangest unsolved mystery?
    \item \textbf{Reference Summary:} The replies present various enigmatic cases, including the construction of the pyramids, unexplained disappearances such as Dyatlov Pass, Bermuda Triangle, and US Navy Flight 19 incidents, as well as more mundane yet mystifying occurrences like a missing TV remote.
    \item \textbf{Model Response:} The disappearance of Amelia Earhart in 1937 remains one of history's most perplexing unsolved mysteries.
\end{itemize}

\begin{table}[H]
\centering
\small
\begin{tabular}{lc}
\toprule
\textbf{Metric} & \textbf{Score} \\
\midrule
BLEU-1 & 0.27\% \\
ROUGE-L F1 & 7.92\% \\
BERTScore F1 & 82.30\% \\
Cosine Similarity & 99.52\% \\
\bottomrule
\end{tabular}
\end{table}

\textbf{Analysis:} The model selects one specific mystery (Amelia Earhart) not mentioned in the reference, yet achieves 99.52\% cosine similarity. Zero n-gram overlap coexists with near-perfect semantic alignment---the model correctly identifies ``unsolved mystery'' as the core concept and provides a valid instance.

\vspace{1em}

\textbf{Example 2: Relationship Advice (Llama-3.1-8B)}

\begin{itemize}[nosep,leftmargin=*]
    \item \textbf{Question:} What's the most real relationship advice you can give?
    \item \textbf{Reference Summary:} Key themes include: open and honest communication; trusting gut instincts; recognizing that love alone is not enough---compatibility, respect, and shared values are essential; letting go of ego; fostering emotional maturity and self-awareness.
    \item \textbf{Model Response:} Communicate openly and honestly with your partner, but also know when to listen without responding. No one is perfect, so accept each other's flaws and work through issues together with empathy and understanding.
\end{itemize}

\begin{table}[H]
\centering
\small
\begin{tabular}{lc}
\toprule
\textbf{Metric} & \textbf{Score} \\
\midrule
BLEU-1 & 0.92\% \\
ROUGE-L F1 & 10.17\% \\
BERTScore F1 & 83.36\% \\
Cosine Similarity & 99.53\% \\
\bottomrule
\end{tabular}
\end{table}

\textbf{Analysis:} The model distills five bullet-pointed themes into a single coherent sentence capturing the core concepts. The 0.92\% BLEU-1 reflects word choice differences (``open communication'' $\rightarrow$ ``communicate openly''), while 99.53\% cosine similarity confirms semantic preservation.

\subsubsection{Edge Cases: Extractive Responses}

Some questions elicit higher lexical overlap when direct content borrowing is appropriate.

\textbf{Example 3: Shower Preferences (Mistral-7B)}

\begin{itemize}[nosep,leftmargin=*]
    \item \textbf{Question:} If you shower once a day, is it in the morning or at night? And how do you feel about that?
    \item \textbf{Reference Summary:} Replies highlighted varied preferences for either morning or evening showers, with some citing benefits such as feeling refreshed, relaxed, or clean, while others mentioned practical considerations like work schedules or exercise routines.
    \item \textbf{Model Response:} Showering once a day can be either in the morning or at night. It's a personal preference and neither is inherently better than the other.
\end{itemize}

\begin{table}[H]
\centering
\small
\begin{tabular}{lc}
\toprule
\textbf{Metric} & \textbf{Score} \\
\midrule
BLEU-1 & 22.43\% \\
ROUGE-L F1 & 30.65\% \\
BERTScore F1 & 88.70\% \\
Cosine Similarity & 99.70\% \\
\bottomrule
\end{tabular}
\end{table}

\textbf{Analysis:} This extractive example shows BLEU-1 at 22.43\% (37$\times$ Mistral's mean), indicating substantial lexical overlap. Such extractive cases constitute $<$2\% of responses, confirming that paraphrasing is the dominant strategy.

\subsubsection{Edge Cases: Variance-Induced Underperformance}

\textbf{Example 4: Workplace Vocabulary (GPT-OSS-20B)}

\begin{itemize}[nosep,leftmargin=*]
    \item \textbf{Question:} What are the most common words you say at work without naming your job?
    \item \textbf{Reference Summary:} Replies reflect a variety of responses, with some sharing industry-specific jargon, others contributing colloquialisms, and a few including humorous comments about workplace vocabulary.
    \item \textbf{Model Response:} Hello, thanks, please, can you, need, update, schedule, project, report, meeting, deadline, confirm, idea, team, feedback, next, steps, issue, solution, progress, review, discuss, sorry.
\end{itemize}

\begin{table}[H]
\centering
\small
\begin{tabular}{lc}
\toprule
\textbf{Metric} & \textbf{Score} \\
\midrule
BLEU-1 & 0.00\% \\
ROUGE-L F1 & 2.06\% \\
BERTScore F1 & 79.99\% \\
Cosine Similarity & 97.95\% \\
\bottomrule
\end{tabular}
\end{table}

\textbf{Analysis:} GPT-OSS-20B's list format diverges from the reference's narrative structure. Despite 0\% BLEU and reduced BERTScore (3.3pp below model mean), cosine similarity remains at 97.95\%---demonstrating that even ``underperforming'' responses maintain strong semantic alignment.

\subsection{Complete Metric Statistics}
\label{sec:appendix-statistics}

Tables~\ref{tab:bleu-stats}--\ref{tab:nli-stats} report comprehensive statistics for all evaluation metrics computed over 11,515 question-response pairs per model.

\begin{table*}[h]
\centering
\begin{tabular}{llccccc}
\toprule
\textbf{Model} & \textbf{Metric} & \textbf{Mean} & \textbf{Med.} & \textbf{Std} & \textbf{Min} & \textbf{Max} \\
\midrule
Gemma2-9B & BLEU-1 & 0.57 & 0.05 & 1.41 & 0 & 33.3 \\
 & BLEU-4 & 0.06 & 0.01 & 0.24 & 0 & 7.1 \\
GPT-OSS-20B & BLEU-1 & 3.71 & 2.34 & 4.02 & 0 & 29.8 \\
 & BLEU-4 & 0.29 & 0.16 & 0.54 & 0 & 10.4 \\
Llama-3.1-8B & BLEU-1 & 7.58 & 7.17 & 5.27 & 0 & 78.6 \\
 & BLEU-4 & 0.62 & 0.34 & 1.85 & 0 & 69.3 \\
Mistral-7B & BLEU-1 & 6.17 & 4.86 & 5.38 & 0 & 31.2 \\
 & BLEU-4 & 0.68 & 0.31 & 1.18 & 0 & 14.1 \\
\bottomrule
\end{tabular}
\caption{BLEU score statistics (\%). Low means with standard deviations exceeding means indicate right-skewed distributions clustered near zero.}
\label{tab:bleu-stats}
\end{table*}

\begin{table*}[h]
\centering
\begin{tabular}{llccccc}
\toprule
\textbf{Model} & \textbf{Metric} & \textbf{Mean} & \textbf{Med.} & \textbf{Std} & \textbf{Min} & \textbf{Max} \\
\midrule
Gemma2-9B & ROUGE-1 & 9.13 & 8.22 & 6.57 & 0 & 40.5 \\
 & ROUGE-L & 6.89 & 6.45 & 4.61 & 0 & 33.3 \\
GPT-OSS-20B & ROUGE-1 & 14.28 & 14.14 & 7.80 & 0 & 56.0 \\
 & ROUGE-L & 9.72 & 9.71 & 4.95 & 0 & 48.0 \\
Llama-3.1-8B & ROUGE-1 & 18.81 & 18.69 & 7.93 & 0 & 86.7 \\
 & ROUGE-L & 11.54 & 11.21 & 4.97 & 0 & 86.7 \\
Mistral-7B & ROUGE-1 & 19.97 & 19.82 & 7.83 & 0 & 48.4 \\
 & ROUGE-L & 13.16 & 12.77 & 5.14 & 0 & 37.2 \\
\bottomrule
\end{tabular}
\caption{ROUGE F1 score statistics (\%). Mistral-7B achieves highest scores; Gemma2-9B lowest.}
\label{tab:rouge-stats}
\end{table*}

\begin{table*}[h]
\centering
\begin{tabular}{llccccc}
\toprule
\textbf{Model} & \textbf{Metric} & \textbf{Mean} & \textbf{Med.} & \textbf{Std} & \textbf{Min} & \textbf{Max} \\
\midrule
Gemma2-9B & Precision & 85.55 & 85.45 & 2.70 & 70.6 & 95.5 \\
 & Recall & 81.18 & 81.12 & 1.59 & 76.3 & 89.2 \\
 & F1 & 83.29 & 83.26 & 1.77 & 74.7 & 92.2 \\
GPT-OSS-20B & Precision & 84.48 & 84.65 & 4.81 & 0.0 & 93.5 \\
 & Recall & 82.25 & 82.58 & 4.40 & 0.0 & 89.2 \\
 & F1 & 83.33 & 83.56 & 4.42 & 0.0 & 89.3 \\
Llama-3.1-8B & Precision & 84.96 & 84.86 & 1.91 & 74.9 & 97.8 \\
 & Recall & 83.14 & 83.13 & 1.63 & 77.8 & 97.8 \\
 & F1 & 84.03 & 83.95 & 1.60 & 77.5 & 97.6 \\
Mistral-7B & Precision & 86.20 & 86.22 & 2.05 & 73.3 & 94.0 \\
 & Recall & 83.51 & 83.52 & 1.55 & 78.2 & 89.1 \\
 & F1 & 84.83 & 84.83 & 1.61 & 77.6 & 90.5 \\
\bottomrule
\end{tabular}
\caption{BERTScore statistics (\%, RoBERTa-large). Precision exceeds recall across all models, indicating concise but semantically relevant outputs.}
\label{tab:bertscore-stats}
\end{table*}

\begin{table*}[h]
\centering
\begin{tabular}{lccccc}
\toprule
\textbf{Model} & \textbf{Mean} & \textbf{Med.} & \textbf{Std} & \textbf{Min} & \textbf{Max} \\
\midrule
Gemma2-9B & 99.20 & 99.25 & 0.27 & 97.9 & 99.8 \\
GPT-OSS-20B & 99.10 & 99.40 & 4.90 & 0.0 & 99.8 \\
Llama-3.1-8B & 99.47 & 99.51 & 0.18 & 97.8 & 99.9 \\
Mistral-7B & 99.51 & 99.54 & 0.16 & 98.2 & 99.8 \\
\bottomrule
\end{tabular}
\caption{Cosine similarity statistics (\%, RoBERTa-large). All models exceed 99\% mean similarity. GPT-OSS-20B shows 30$\times$ higher variance than Mistral-7B.}
\label{tab:cosine-stats}
\end{table*}

\begin{table*}[h]
\centering
\begin{tabular}{lccccc}
\hline
\textbf{Model} & \textbf{Mean} & \textbf{Med.} & \textbf{Std} & \textbf{Min} & \textbf{Max} \\
\hline
Gemma2-9B & 50.92 & 50.88 & 2.27 & 43.64 & 60.16 \\
GPT-OSS-20B & 51.84 & 52.04 & 3.42 & 0.00 & 59.38 \\
Llama-3.1-8B & 52.65 & 52.53 & 2.07 & 44.37 & 69.97 \\
Mistral-7B & 53.42 & 53.39 & 2.08 & 44.91 & 61.99 \\
\hline
\end{tabular}
\caption{MoverScore statistics (\%). Mistral-7B achieves the highest mean score. GPT-OSS-20B shows notably higher variance (std = 3.42) with zero-value outliers, consistent with its erratic patterns observed in other metrics. MoverScore's intermediate range (50.9--53.4\%) sits between lexical metrics (<8\% BLEU-1) and other semantic metrics (>83\% BERTScore, >99\% cosine similarity).}
\label{tab:moverscore-stats}
\end{table*}

\begin{table*}[h]
\centering
\begin{tabular}{llccccc}
\toprule
\textbf{Model} & \textbf{Label} & \textbf{Mean} & \textbf{Med.} & \textbf{Std} & \textbf{Min} & \textbf{Max} \\
\midrule
Gemma2-9B & Entailment & 9.04 & 0 & 27.4 & 0 & 100 \\
 & Contradiction & 5.15 & 0 & 21.0 & 0 & 100 \\
 & Neutral & 85.81 & 100 & 33.2 & 0 & 100 \\
GPT-OSS-20B & Entailment & 3.43 & 0 & 15.8 & 0 & 100 \\
 & Contradiction & 6.94 & 0 & 23.7 & 0 & 100 \\
 & Neutral & 89.39 & 100 & 28.0 & 0 & 100 \\
Llama-3.1-8B & Entailment & 4.06 & 0 & 15.1 & 0 & 100 \\
 & Contradiction & 5.53 & 0 & 18.1 & 0 & 100 \\
 & Neutral & 90.41 & 100 & 22.8 & 0 & 100 \\
Mistral-7B & Entailment & 8.71 & 0 & 23.3 & 0 & 100 \\
 & Contradiction & 4.31 & 0 & 16.8 & 0 & 100 \\
 & Neutral & 86.98 & 100 & 27.6 & 0 & 100 \\
\bottomrule
\end{tabular}
\caption{NLI distribution statistics (\%, BART-large-MNLI). Neutral classifications dominate (86--90\%), indicating models generate semantically related but non-entailing content.}
\label{tab:nli-stats}
\end{table*}

\subsection{Distribution Visualizations}
\label{sec:appendix-figures}

This section presents visual comparisons of evaluation metrics across all four models. The figures confirm the semantic-lexical paradox quantitatively: lexical metrics (BLEU, ROUGE) show substantial cross-model variation, semantic metrics (BERTScore, Cosine Similarity) cluster near ceiling values, and MoverScore occupies an intermediate position reflecting the optimal transport cost of semantic alignment.

\subsubsection{Lexical Overlap Metrics}
\begin{figure*}[ht]
\centering
\includegraphics[width=1.8\columnwidth]{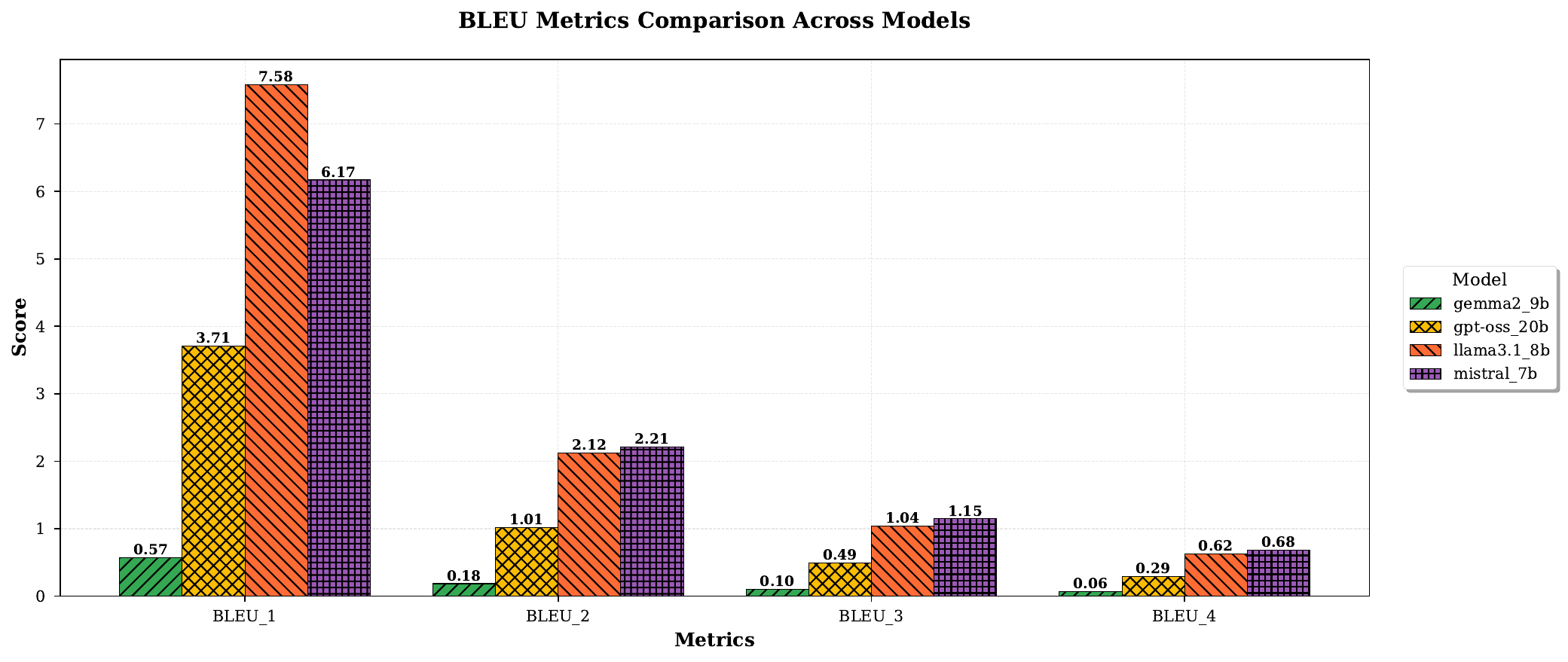}
\caption{BLEU score comparison across models. BLEU-1 ranges from 0.57\% (Gemma2-9B) to 7.58\% (Llama-3.1-8B)---a 13$\times$ difference---while BLEU-4 scores remain below 1\% for all models. Llama-3.1-8B's elevated scores may partially reflect self-alignment bias, as it was also used for reference summarization.}
\label{fig:bleu-comparison}
\end{figure*}

\begin{figure*}[ht]
\centering
\includegraphics[width=1.8\columnwidth]{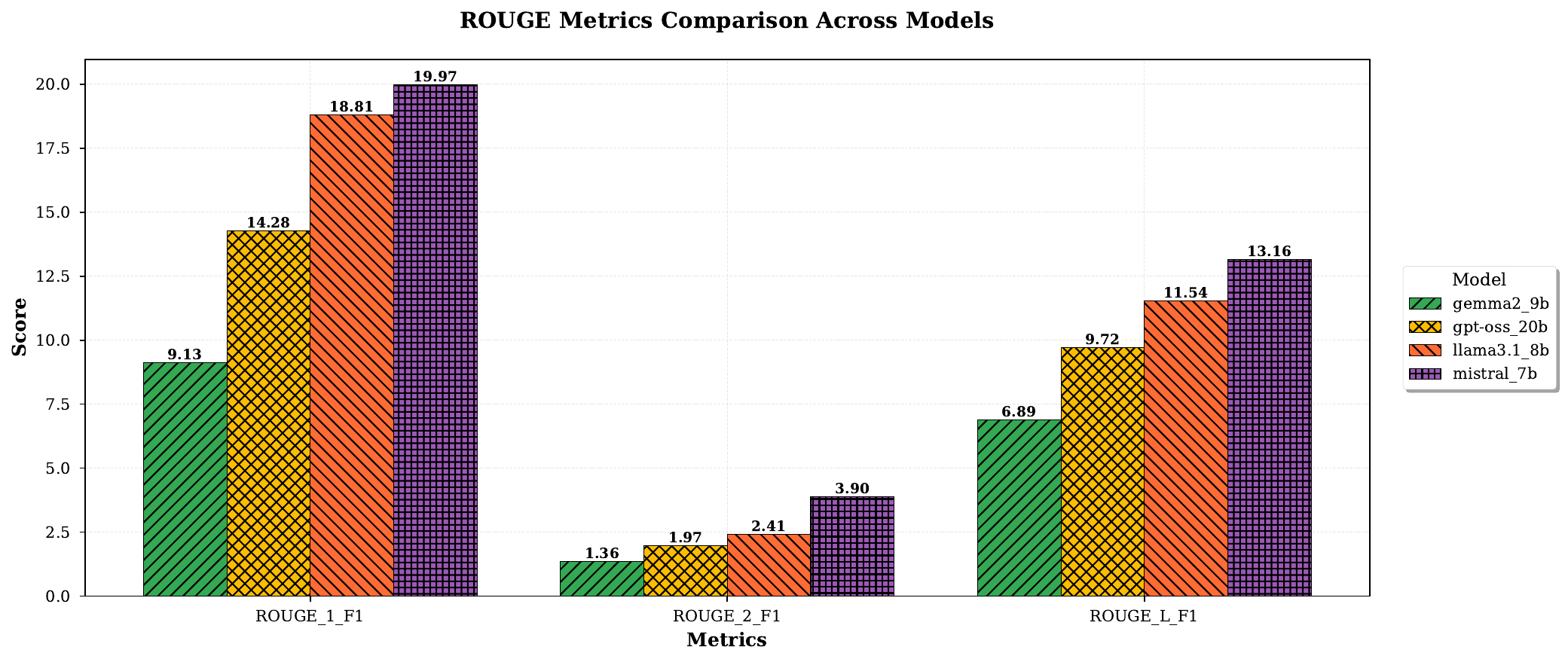}
\caption{ROUGE F1 score comparison across models. Mistral-7B achieves highest scores across all variants (ROUGE-1: 19.97\%, ROUGE-L: 13.16\%), followed closely by Llama-3.1-8B. Gemma2-9B shows lowest overlap (ROUGE-1: 9.13\%), consistent with its extreme abstraction profile. ROUGE-2 scores remain below 4\% for all models, indicating minimal bigram overlap.}
\label{fig:rouge-comparison}
\end{figure*}

Figures~\ref{fig:bleu-comparison} and~\ref{fig:rouge-comparison} show BLEU and ROUGE scores across models. 

\subsubsection{Semantic Similarity Metrics}
\begin{figure*}[ht]
\centering
\includegraphics[width=1.8\columnwidth]{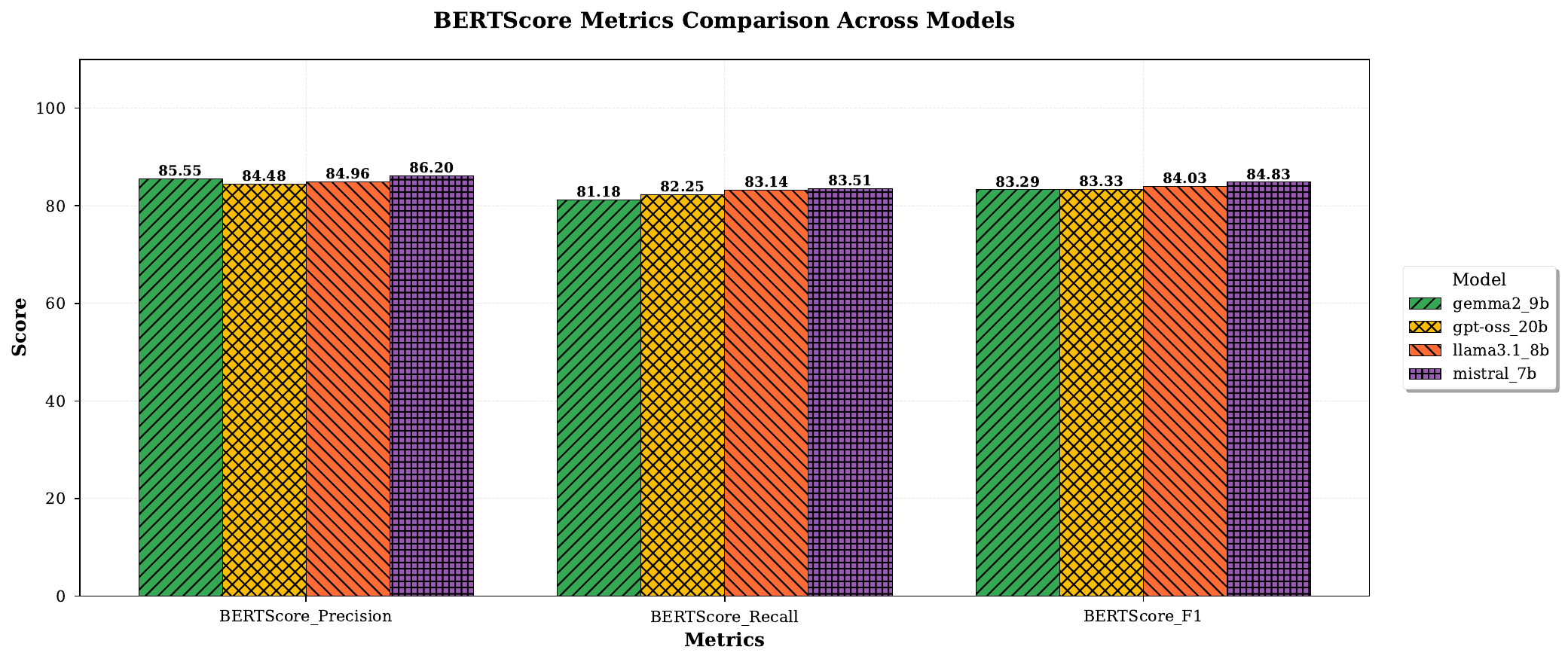}
\caption{BERTScore comparison across models (RoBERTa-large embeddings). Despite the 13$\times$ variation in BLEU-1 scores (Figure~\ref{fig:bleu-comparison}), BERTScore F1 varies by only 1.54 percentage points (83.29\%--84.83\%). Precision exceeds recall for all models, indicating generated answers contain semantically relevant content but are more concise than reference summaries.}
\label{fig:bertscore-comparison}
\end{figure*}

\begin{figure*}[ht]
\centering
\includegraphics[width=1.5\columnwidth]{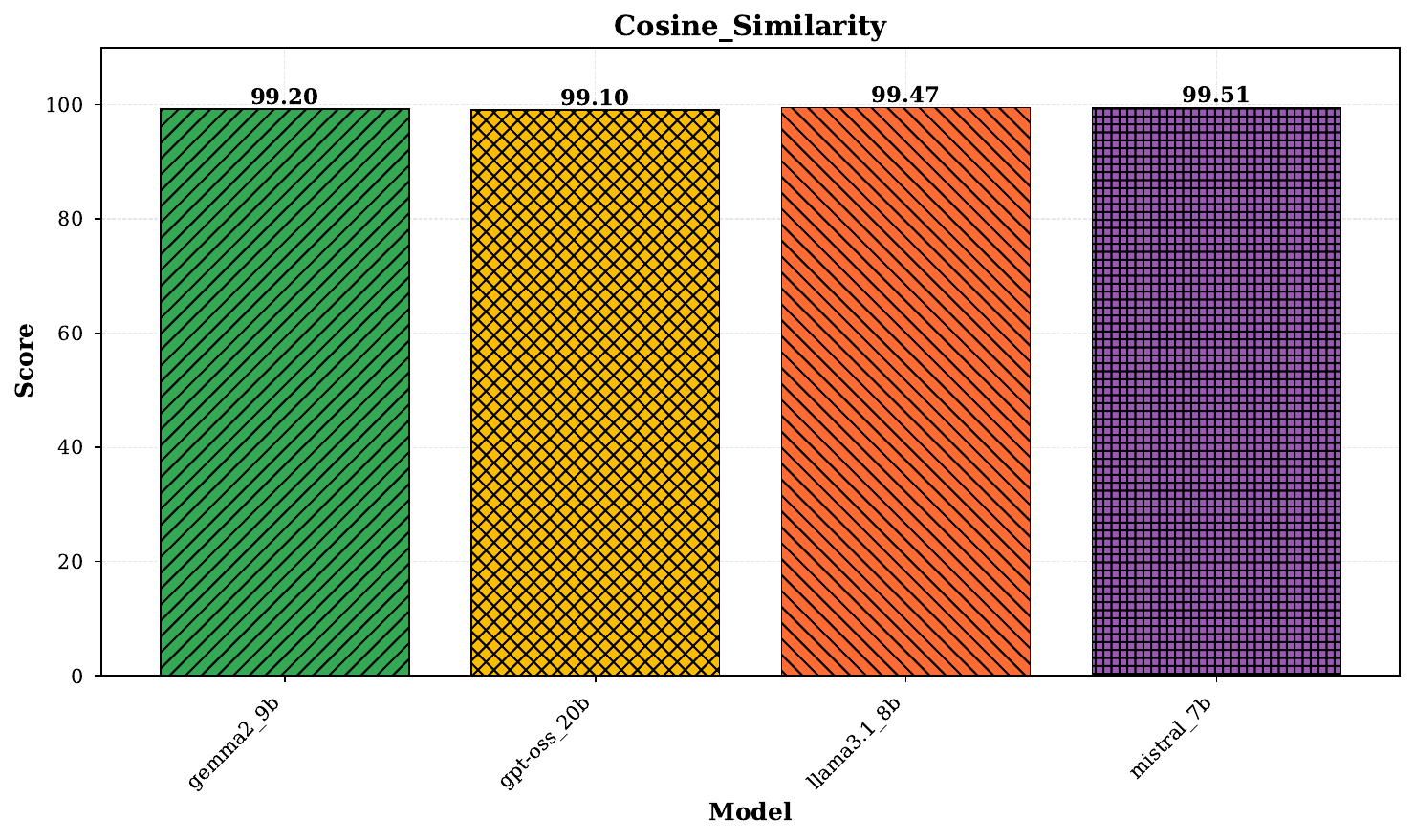}
\caption{Cosine similarity comparison across models. All models exceed 99\% similarity in RoBERTa-large embedding space, with only 0.41 percentage points separating Mistral-7B (99.51\%) from GPT-OSS-20B (99.10\%). This near-ceiling performance contrasts sharply with the $<$8\% BLEU-1 scores---the 90+ percentage point gap constitutes our central finding.}
\label{fig:cosine-comparison}
\end{figure*}

Figures~\ref{fig:bertscore-comparison} and~\ref{fig:cosine-comparison} show BERTScore and cosine similarity and MoverScore.

\begin{figure*}[ht]
\centering
\includegraphics[width=1.5\columnwidth]{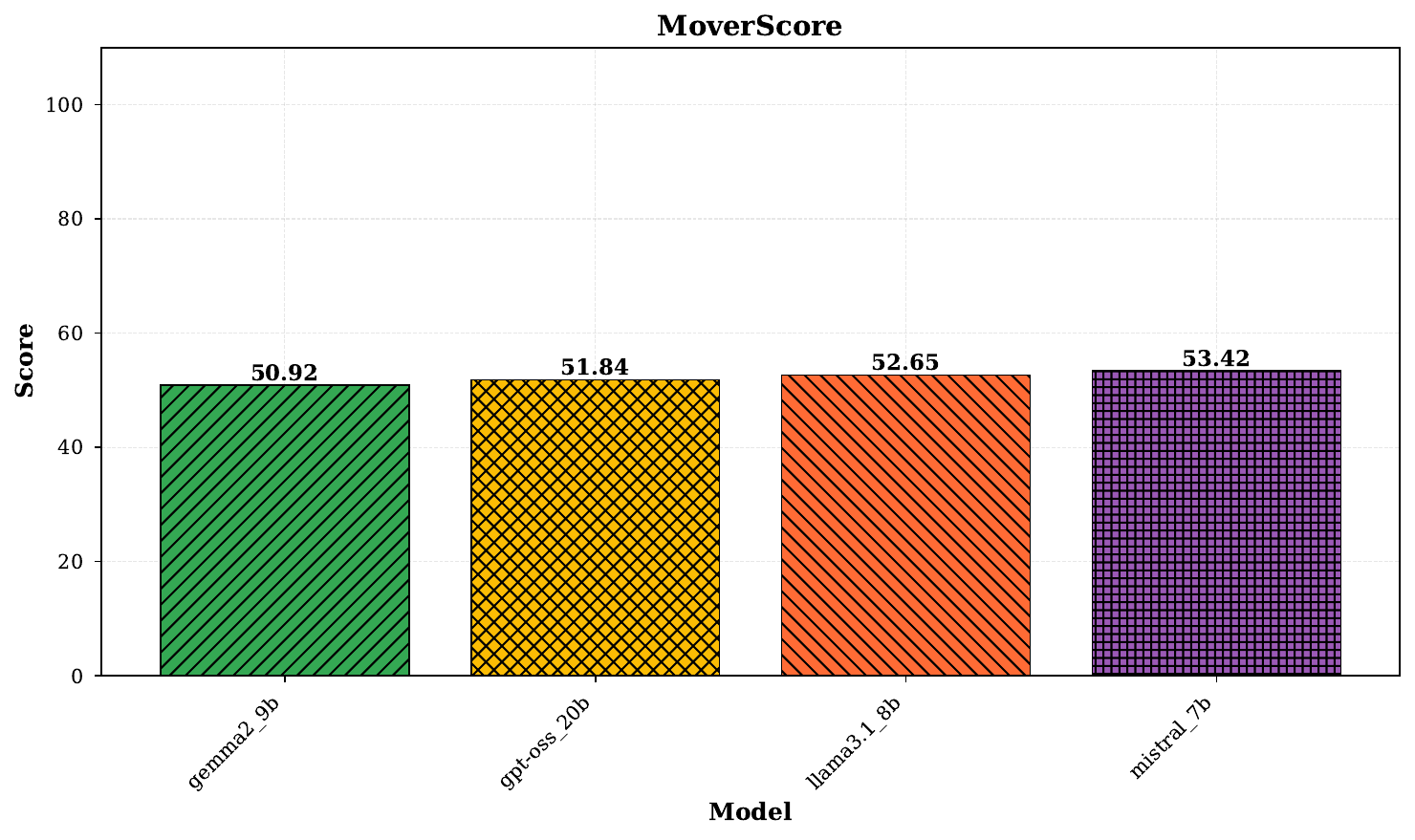}
\caption{MoverScore comparison across models. Mistral-7B achieves the highest score (53.42\%), followed by Llama-3.1-8B (52.65\%), GPT-OSS-20B (51.84\%), and Gemma2-9B (50.92\%).}
\label{fig:moverscore}
\end{figure*}

\subsubsection{Logical Inference Patterns}
\begin{figure*}[ht]
\centering
\includegraphics[width=1.5\columnwidth]{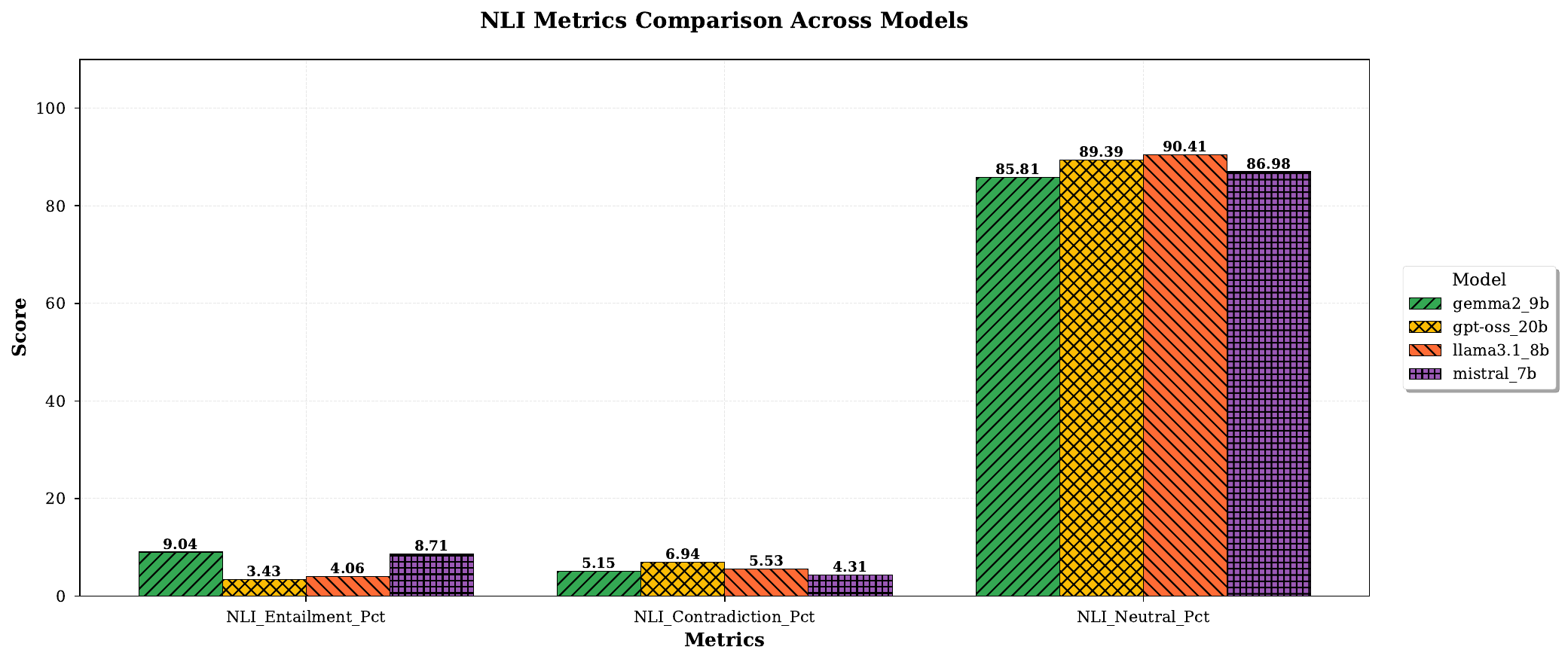}
\caption{NLI classification distribution across models. Neutral classifications dominate (86--90\%), with entailment rates varying from 3.43\% (GPT-OSS-20B) to 9.04\% (Gemma2-9B). Contradiction rates remain below 7\% for all models, suggesting direct factual conflicts with human consensus are rare. See Table~\ref{tab:nli-results} in Section 4.4 for exact values.}
\label{fig:nli-comparison}
\end{figure*}

Figure~\ref{fig:nli-comparison} shows NLI classification distributions.

\subsection{Model Behavior Summary}
\label{sec:appendix-models}

Our analysis reveals three key patterns:

\textbf{Abstraction-verbosity spectrum:} Models range from extreme abstraction (Gemma2-9B) to extractive-verbose (Llama-3.1-8B). Despite a 13$\times$ difference in lexical overlap, cosine similarity varies by less than 0.5 percentage points.

\textbf{Consistency:} Mistral-7B is 30$\times$ more consistent than GPT-OSS-20B, suggesting architectural or training differences in handling diverse question types.

\textbf{Scale $\neq$ performance:} GPT-OSS-20B (20B parameters) underperforms Mistral-7B (7B parameters) across all metrics.

\end{document}